\documentclass[journal]{IEEEtran}
\usepackage{times}  
\usepackage{helvet} 
\usepackage{courier}  
\usepackage[hyphens]{url}  
\usepackage{graphicx} 
\usepackage{graphicx}  
\usepackage{booktabs}
\usepackage{subfigure}
\usepackage{amsmath}
\usepackage{amsfonts}
\usepackage{verbatim}
\usepackage{multirow}
\usepackage{multicol}

\begin{document}

\title{Information Symmetry Matters: A Modal-Alternating Propagation Network for Few-Shot Learning}

\author{Zhong~Ji,~\IEEEmembership{Senior~Member,~IEEE,}
		Zhishen~Hou,
        Xiyao~Liu,~\IEEEmembership{Graduate~Student~Member,~IEEE,}
        Yanwei~Pang,~\IEEEmembership{Senior~Member,~IEEE,}
        Jungong~Han,~\IEEEmembership{Senior~Member,~IEEE}

\thanks{Manuscript received xxx xx, 2021; revised xxx xx, 2021.}
\thanks{This work was supported by the National Natural Science Foundation of China (NSFC) under Grants 61771329 and 61632018.}
\thanks{Z. Ji, Z. Hou, X. Liu*(corresponding author), and Y. Pang are with the School of Electrical and Information Engineering, Tianjin University, Tianjin 300072, China (e-mails: jizhong@tju.edu.cn; zshou@tju.edu.cn; xiyaoliu@tju.edu.cn; pyw@tju.edu.cn).} 
\thanks{J. Han is with the Computer Science Department, Aberystwyth University, Aberystwyth SY23 3FL, U.K. (e-mail: jungong.han@aber.ac.uk).}}

\markboth{Journal of IEEE TRANSACTIONS ON IMAGE PROCESSING,~Vol.~nn, No.~n, August~20nn}%
{Shell \MakeLowercase{\textit{et al.}}: Bare Demo of IEEEtran.cls for IEEE Journals}

\maketitle

\begin{abstract}
Semantic information provides intra-class consistency and inter-class discriminability beyond visual concepts, which has been employed in Few-Shot Learning (FSL) to achieve further gains. However, semantic information is only available for labeled samples but absent for unlabeled samples, in which the embeddings are rectified unilaterally by guiding the few labeled samples with semantics. Therefore, it is inevitable to bring a cross-modal bias between semantic-guided samples and nonsemantic-guided samples, which results in an information asymmetry problem. To address this problem, we propose a Modal-Alternating Propagation Network (MAP-Net) to supplement the absent semantic information of unlabeled samples, which builds information symmetry among all samples in both visual and semantic modalities. Specifically, the MAP-Net transfers the neighbor information by the graph propagation to generate the pseudo-semantics for unlabeled samples guided by the completed visual relationships and rectify the feature embeddings. In addition, due to the large discrepancy between visual and semantic modalities, we design a Relation Guidance (RG) strategy to guide the visual relation vectors via semantics so that the propagated information is more beneficial. Extensive experimental results on three semantic-labeled datasets, i.e., Caltech-UCSD-Birds 200-2011, SUN Attribute Database and Oxford 102 Flower, have demonstrated that our proposed method achieves promising performance and outperforms the state-of-the-art approaches, which indicates the necessity of information symmetry.
\end{abstract}


%
\IEEEpeerreviewmaketitle

\section{Introduction}
\IEEEPARstart{D}{eep} learning has become the dominant technology on computer vision tasks. However, the performance is severely limited by the amount of labeled data, which is difficult or even infeasible to be acquired due to the high annotation cost or the scarcity of rare categories. By contrast, humans can recognize a new object with only one or limited observations based on abundant prior knowledge learned before. Inspired by this, learning to recognize new classes with few samples, called Few-Shot Learning (FSL) \cite{chen2019a,wang2021trust}, has attracted great attention recently.

\begin{figure}[t]
	\begin{center}
		\includegraphics[width=0.98\linewidth]{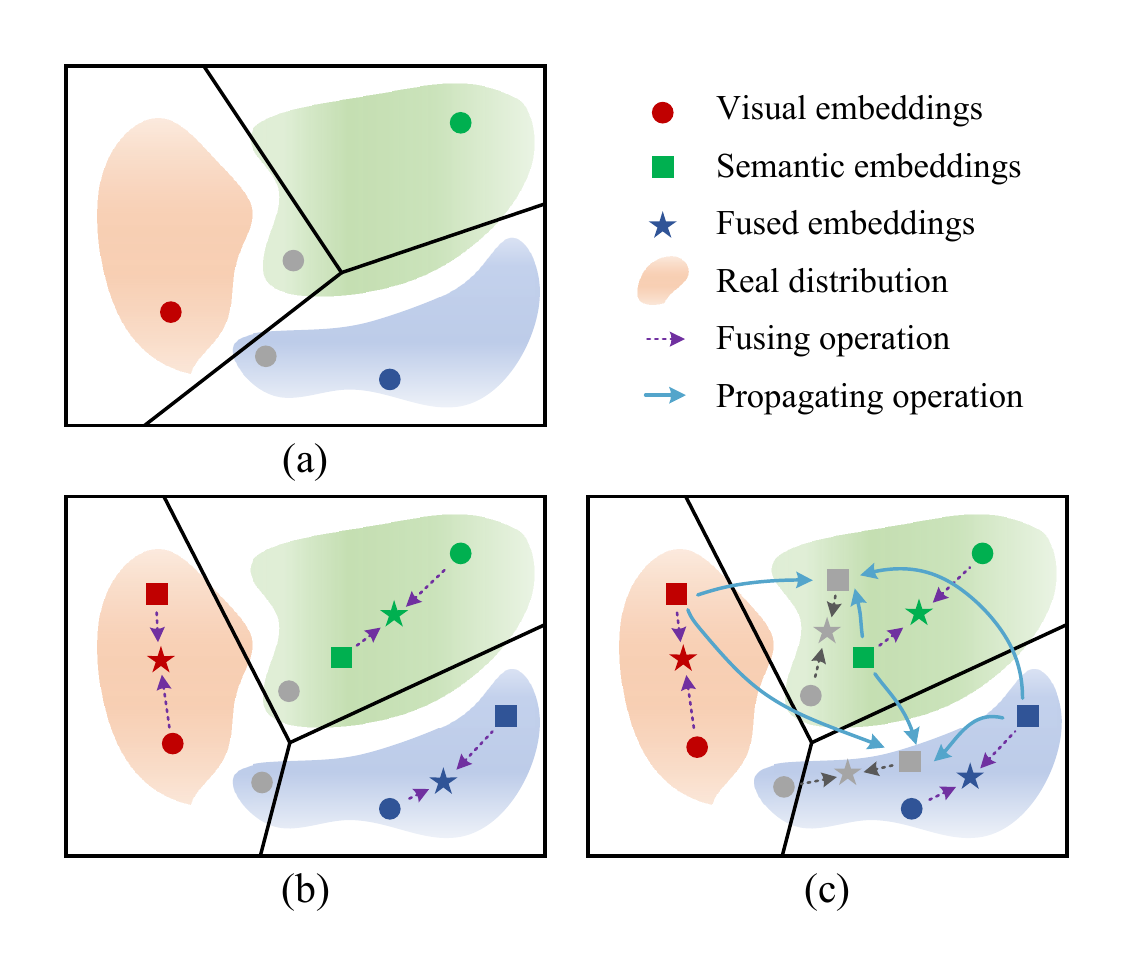}
	\end{center}
	\label{fig1}
	\caption{ An illustration of information asymmetries between support set and query set in FSL. (a) Embeddings with only visual features; (b) Support embeddings fused with visual features and semantic vectors and query embeddings with only visual features; (c) Query embeddings fused with visual features and propagated pseudo-semantic vectors.}
\end{figure}


One intuitive solution for FSL is to employ the experience learned from other similar tasks. Meta-learning \cite{hospedales2021meta,zhu2020personalized}, also known as “learning to learn”, aims at learning new concepts or skills rapidly with a few training examples based on abundant prior knowledge learned from base classes. Thus, the meta-learning framework has been widely employed on FSL and achieved promising performance \cite{vinyals2016matching,Snell2017Prototypical,sung2018learning}.

Many studies demonstrate that humans capture object concepts from not only the visual view but also the language describing the characteristics of the objects \cite{jackendoff1987beyond,wallis1999learning}. Thus, some FSL approaches \cite{xing_adaptive_2019,chen2019multi,li_boosting_2020,huang_attributes-guided_2021} utilize auxiliary semantic information, i.e., word embeddings or attribute annotations, to enhance the feature representations and improve the performance. For example, Li $et\, al.$ \cite{li_boosting_2020} modified the visual embeddings according to the relationship between the visual distance and the semantic similarity of different categories. Huang $et\, al.$ \cite{huang_attributes-guided_2021} employed an attribute-guided attention mechanism to augment the representations with the guidance of semantics. However, since the semantic information of query samples is unavailable, the utilization of query embeddings and the support embeddings enhanced with semantics will inevitably produce a cross-modal embedding bias, which may lead to an information asymmetry problem, as shown in Figure 1(b).

To address this problem, we introduce the graph propagation model to obtain the query semantics by the completed relation information in visual modality. The graph structure \cite{wang2021hash} has natural advantages on modeling relationships among nodes, which is effective to propagate information from one node to another. The main idea of our method is to update the visual graph with the guidance of semantics, and then propagate the semantic graph with the information transferred from visual modality. With the alternating propagation in two modalities, the information asymmetry problem is alleviated significantly as the visual graph is rectified and the semantic graph is completed.

Considering the large discrepancy across modalities, it is essential to reduce the cross-modal shift so that the information in two modalities is beneficial for each other. A common approach is to constrain the embeddings of two modalities in a shared latent space with a penalty function. For example, Schonfeld $et\, al.$ \cite{schonfeld2019generalized} learned shared cross-modal feature embeddings of visual and semantic modalities with a Variational Auto-Encoder (VAE) and aligned the embeddings with two elaborate loss functions. Tokmakov $et\, al.$ \cite{tokmakov_learning_2019} designed a soft constraint regularization which improves the robustness of the alignments. However, directly constraining the instance embeddings is inappropriate in FSL since it is difficult to maintain a balance between extracting discriminative features and aligning cross-modal embeddings in a low-data scenario. To this end, we focus on the relations among samples and design a new guidance strategy which is flexible to reduce the cross-modal discrepancy.

Specifically, we propose a Modal-Alternating Propagation Network (MAP-Net) to obtain semantic information of query samples and rectify the feature embeddings, which alleviates the information asymmetry problem in FSL. The MAP-Net constructs two graphs in two modalities, i.e., the visual graph and the semantic graph, which are propagated alternately with the guidance of each other modality. The semantic graph is incomplete since semantic embeddings of query samples are unavailable. Therefore, we transfer the relation information from visual modality to semantic modality which is essential to propagate and complete the semantic graph. After the propagation, the information asymmetries are mitigated significantly as the query semantics are generated. To reduce the cross-modal discrepancy, we propose a Relation Guidance (RG) strategy to modify the relationships in visual modality. We transfer the relation vectors with a relation transfer module which is trained with support-support pairs to obtain the rectified relationships.

Our highlights are summarized in three folds:

\begin{itemize}
	\item We propose a Modal-Alternating Propagation Network to propagate modal information alternately to generate the pseudo-semantics of query samples and rectify the feature embeddings, which is effective in alleviating the information asymmetry problem between support and query samples.

	\item To overcome the discrepancy between modalities and obtain the accurate relation information among different samples, we propose a Relation Guidance strategy to guide visual relationships with the relationships in semantic modality. The visual relation vectors are transferred with a relation transfer module trained with support-support pairs to represent the relationships more accurately. 

	\item We conduct experiments on three benchmark datasets with attributes or text descriptions, i.e., Caltech-UCSD-Birds 200-2011, SUN Attribute Database and Oxford 102 Flower, to compare the proposed method with previous few-shot learning methods. The experimental results demonstrate that our method achieves promising performance for few-shot learning from the perspective of information symmetry. 
\end{itemize}

\begin{figure*}[t]
	\tiny
	\begin{center}
		\includegraphics[width=\textwidth]{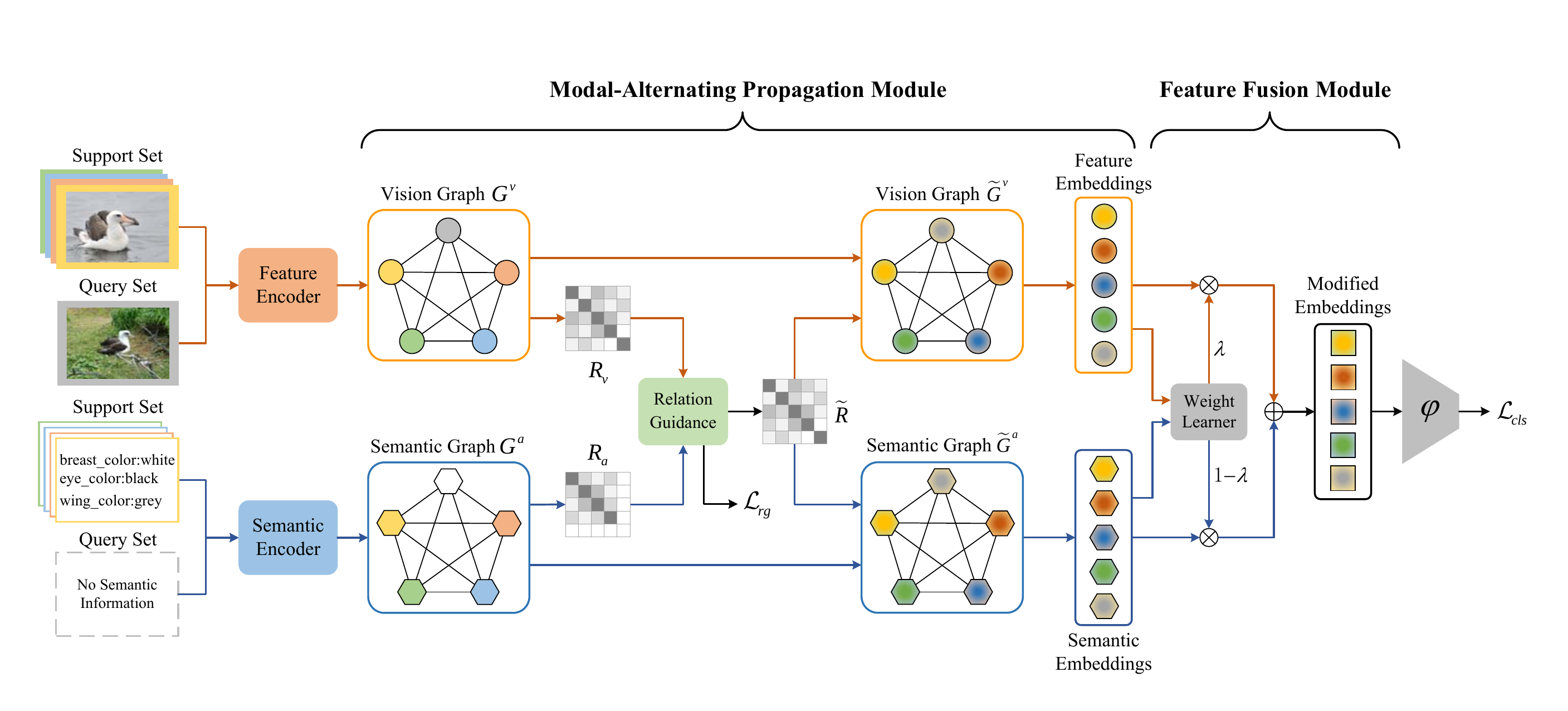}
	\end{center}
	\label{fig2}
	\caption{The framework of MAP-Net for the 4-way 1-shot task. Semantic information of support set is given, while it is unknown in query set. The pseudo-semantic embeddings of query samples are generated by semantic graph in the MAP-Module, and the output embeddings in two modalities are fused for classification.
	}
\end{figure*}

\section{Related work}
\subsection{Few-Shot Learning}
Few-Shot Learning aims at learning novel concepts with only one or few objects, which has been widely studied in recent years. Most of the existing few-shot methods follow the meta-learning strategy, which is also known as learning to learn, to transfer prior knowledge obtained from a large amount of auxiliary data in the meta-training phase to the novel tasks. The meta-learning-based methods generally can be divided into three types: optimization-based methods, metric-based methods and data-augmentation-based methods.

The optimization-based methods learn sub-optimal parameters for every task as the initial parameters that can be quickly adapted to novel tasks by only a few steps of gradient descent. MAML \cite{finn2017model-agnostic} is the first optimization-based method that utilizes a second-order optimizing strategy with meta-learning framework to quickly update the parameters. To simplify the optimization, Nichol $et\, al.$ \cite{nichol2018first} utilized the first-order method to replace the second-order derivation in MAML. Rusu $et\, al.$ \cite{rusu_meta-learning_2019} updated parameters in a low-dimensional latent space which is more practical in low-data scenarios. Since the shared initialization may lead to the conflict over tasks, Baik $et\, al.$ \cite{baik_learning_2020} proposed a task-and-layer-wise attenuation to forget the prior information selectively.

In the metric-based methods, the embedding space is constructed to measure the similarities among different feature embeddings. The simplicity and efficiency make metric-based methods highly attractive in the field of few-shot learning. Matching Networks \cite{vinyals2016matching} utilize an attention mechanism based on LSTM to learn to classify the novel samples. Snell $et\, al.$ \cite{Snell2017Prototypical} proposed Prototypical Networks to measure the distance between each sample and the prototypes of corresponding class. Relation Network \cite{sung2018learning} utilizes a learnable metric method instead of manual measurement, which is more flexible in metric-based classifications.

The main idea of the data-augmentation-based methods is to alleviate the lack of labeled data with data augmentation. Wang $et\, al.$ \cite{wang2018low-shot} generated samples with the idea of GAN to expand the diversity of data. Zhang $et\, al.$ \cite{zhang2019few-shot} utilized a saliency detection method to fuse foreground and background in different images to augment samples. In order to alleviate the mode collapse problem in GAN, Li $et\, al.$ \cite{li2020adversarial} employed the cWGAN in few-shot learning to ensure the diversity of generated data.

\subsection{Learning with Semantic Information}
Semantic information usually plays a crucial role in various tasks, such as image-text matching \cite{wang2020consensus,Chen_2020_CVPR}, emotion recognition \cite{kosti2017emotion,zhang2019spatial} and zero-shot learning \cite{frome2013devise,ji2020attribute}. When the samples in visual modality are scarce, the semantic information is a good choice to assist model in training. Many zero-shot learning methods align visual and semantic representation to achieve the classification of novel classes without labeled samples. For example, Frome $et\, al.$ \cite{frome2013devise} presented a deep visual-semantic embedding model to learn the semantic relationships among classes with the semantic data, and map the visual samples into a semantic space to be classified. Ji $et\, al.$ \cite{ji2020attribute} employed a semantic embedding space to transfer knowledge from seen domain to unseen domain with an attribute-guided network to address the cross-modal zero-shot hashing retrieval tasks. Considering the potential bias between seen and unseen classes, Gao $et\, al.$ \cite{gao2020zero} utilized a joint generative model to generate high-quality unseen features, which is further augmented with a self-training strategy. Guan $et\, al.$ \cite{guan2020zero} learned a robust cross-modal projection by synthesizing the unseen class data and designing a novel projection learning model to best utilize the synthesized data. 


Based on the success of zero-shot learning, some methods utilizing auxiliary semantic information are proposed to boost the few-shot learning in recent years. Xing $et\, al.$ \cite{xing_adaptive_2019} integrated the semantic embeddings into visual features with an adaptive convex combination to assist the classification. Similarly, SAP-Net \cite{ji2021hoi} refines the embeddings with the guidance of semantic information. Schwartz $et\, al.$ \cite{schwartz_baby_2019} further improved the few-shot learning with multiple semantics. For the discrepancy between visual and semantic modalities, Tokmakov $et\, al.$ \cite{tokmakov_learning_2019} designed a semantic-based soft constraint regularization to learn the compositional representation of each sample. Chen $et\, al.$ \cite{chen2019multi} synthesized sample features in a semantic space with an encoder-decoder framework to increase the diversity of feature embeddings. 
Huang $et\, al.$ \cite{huang_attributes-guided_2021} employed the attention mechanism to emphasize or suppress the representation with the guidance of semantic information. Zhang $et\, al.$ \cite{zhang_rethinking_2021} addressed few-shot classification in both relative and absolute views, which utilizes the semantic information and class labels simultaneously to represent the similarities among different samples and the absolute concept of instances.

\subsection{Propagation with Graph Model}
Graph model is effective in constructing the relationship among different nodes, which has received great attention in recent years. Thanks to the great expressive power of graphs, especially the convincing performance of deep learning based Graph Neural Network (GNN) \cite{garcia2018fewshot}, the graph-based methods have been employed in types of tasks, e.g., nodes classification \cite{kim_edge-labeling_2019}, link prediction \cite{zhang2018link}, and clustering \cite{zhang2019heterogeneous}. 

In this case, various of graph-based methods are proposed for few-shot learning to extract the relationship and rectify the graph in an episode. In order to further explore the potential of graph model in few-shot learning, Kim $et\, al.$ \cite{kim_edge-labeling_2019} proposed EGNN to dynamically update both the nodes and edges and Yang $et\, al.$ \cite{yang_dpgn_2020} constructed both the distribution-level relations and instance-level relations with DPGN. Moreover, Label Propagation (LP) \cite{zhou2004learning} is also a classical method to transfer knowledge from neighbors of each node, which is proved effective for few-shot classification with meta-learning framework. TPN \cite{liu2019learning} is the first method utilizing label propagation in few-shot learning. After this, Rodriguez $et\, al.$ \cite{rodriguez_embedding_2020} employed an Embedding Propagation method to yield a smoother embedding manifold. Similarly in zero-shot learning, Liu $et\, al.$ \cite{liu2020attribute} optimized the semantic space with Attribute Propagation Network to refine the attributes of each class. Inspired by this, our method aims at propagating information both in visual and semantic space with cross-modal assistance, called Modal-Alternating Propagation (MAP-Net).

\section{Methodology}
\subsection{Preliminary}

We follow the episodic training paradigm as \cite{vinyals2016matching} for few-shot learning. In general, our model is trained on $N$-way $K$-shot settings. Each episode consists of $N$ categories from meta-training set and is divided into two parts: $K$ labeled samples in the support set and several query samples in the query set. Thus, the semantic information of support set is available. The support set contains totally $N \times K$ labeled samples with their semantic vectors. It is denoted as $S=\{ x_i^s, a_i^s, y_i^s \}_{i=1}^{N \times K}$, where $x_i, a_i, y_i$ represents the $i$-th image, semantic vectors and the corresponding label. The query set $Q=\{ x_i^q, y_i^q \}_{i=1}^{T}$ contains $T$ samples with no semantic vectors. The episodic paradigm aims at obtaining the optimal performance on the query set by training the model with the support set.

\subsection{Overview}

In this work, we propose a Modal-Alternating Propagation Network (MAP-Net) for few-shot learning to rectify the feature embeddings by constructing information symmetry. Figure 2 presents the main framework of MAP-Net, which consists of a modal-alternating propagation module (MAP-Module) and a feature fusion module. In the MAP-Module, two graphs in both visual and semantic modalities are constructed for learning to propagate information alternately to obtain the semantic vectors of query samples. Concretely, we first modify the visual embeddings with the guidance of semantic to reduce the high intra-class variance in visual modality. Then the semantic graph will be completed by applying the information transferred from visual space. We employ a Relation Guidance (RG) strategy to guide the visual embeddings. Rather than simply aligning two modal embeddings with a penalty function, we guide the relation information among samples in visual modality with that in semantic modality. With the guidance of relation map, both the visual graph and semantic graph are updated to obtain the symmetrical embeddings of support and query samples. Finally, the propagated visual and semantic embeddings are fused as augmented embeddings by a convex combination as \cite{xing_adaptive_2019}, which are employed to classify in corresponding categories.

\subsection{Modal-Alternating Propagation}

Since the categories of query samples are unknown during test stage, their semantic information is unavailable. This may lead to an information asymmetry between support samples and query samples. To reduce the bias caused by the lack of the query semantics, we design a Modal-Alternating Propagation Module (MAP-Module) to generate the pseudo-semantic embeddings through updating graphs with propagating operation.

\textbf{Graph Construction}. There are two types of graphs, the visual graph $G^v=(V^v,E^v)$ and the semantic graph $G^a=(V^a,E^a)$ in MAP-Module. They transfer and propagate information to generate the pseudo query semantics. In the visual graph $G^v$, the node $V_i^v$ is the feature embedding $z_i^v \in \mathbb{R}^{c}$ obtained from a feature encoder (CNN), which can be denoted as

\begin{equation}
	z_i^v = f(x_i),
\end{equation}
where $f$ represents the feature encoder. For semantic graph, we utilize the attributes or text descriptions of samples as semantic information. Each node $V_i^a$ is also an embedding $z_i^a \in \mathbb{R}^c$ of semantic information encoded by a semantic encoder. The difference is the query semantic embeddings are initialized with zero vectors. $z^a$ is defined as:

\begin{equation}
	z_i^a = \left\{
		\begin{aligned}
			& g(a_i) &, &  a_i \in S, \\
			& 0 &, &  a_i \in Q,
		\end{aligned}
	\right.
\end{equation}
where $g$ is a multi-layer perceptron (MLP) used to encode the semantic vectors. Then the adjacency matrix $A^v$ in $G^v$ is obtained from the similarities among nodes. In this paper, we employ Gaussian similarity function $A_{ij}^v = exp (-d^2 (z_i,z_j) / \sigma^2 )$
to calculate graph edges, where $d$ is the Euclidean distance between two neighbor nodes and $\sigma$ is a scaling factor. Especially, we make the values of diagonal $A_{ii}^v=0$ to avoid self-reinforcement. In this paper, we utilize the standard deviation of distance matrix $std(d^2)$ as $\sigma^2$ as in \cite{rodriguez_embedding_2020}.

The adjacency matrix of semantic graph $A^a$ is similar as $A^v$. However, the query semantic vectors are inapplicable to provide the relationships among different nodes in semantic graph. To address this problem, we explore to transfer the relation information from the visual graph to the semantic graph to acquire the completed adjacency matrix $\widetilde{A}^a$, which can be applied to propagate information to generate the pseudo-semantic embeddings of query samples.

\begin{figure}[t]
	\begin{center}
		\includegraphics[width=0.98\linewidth]{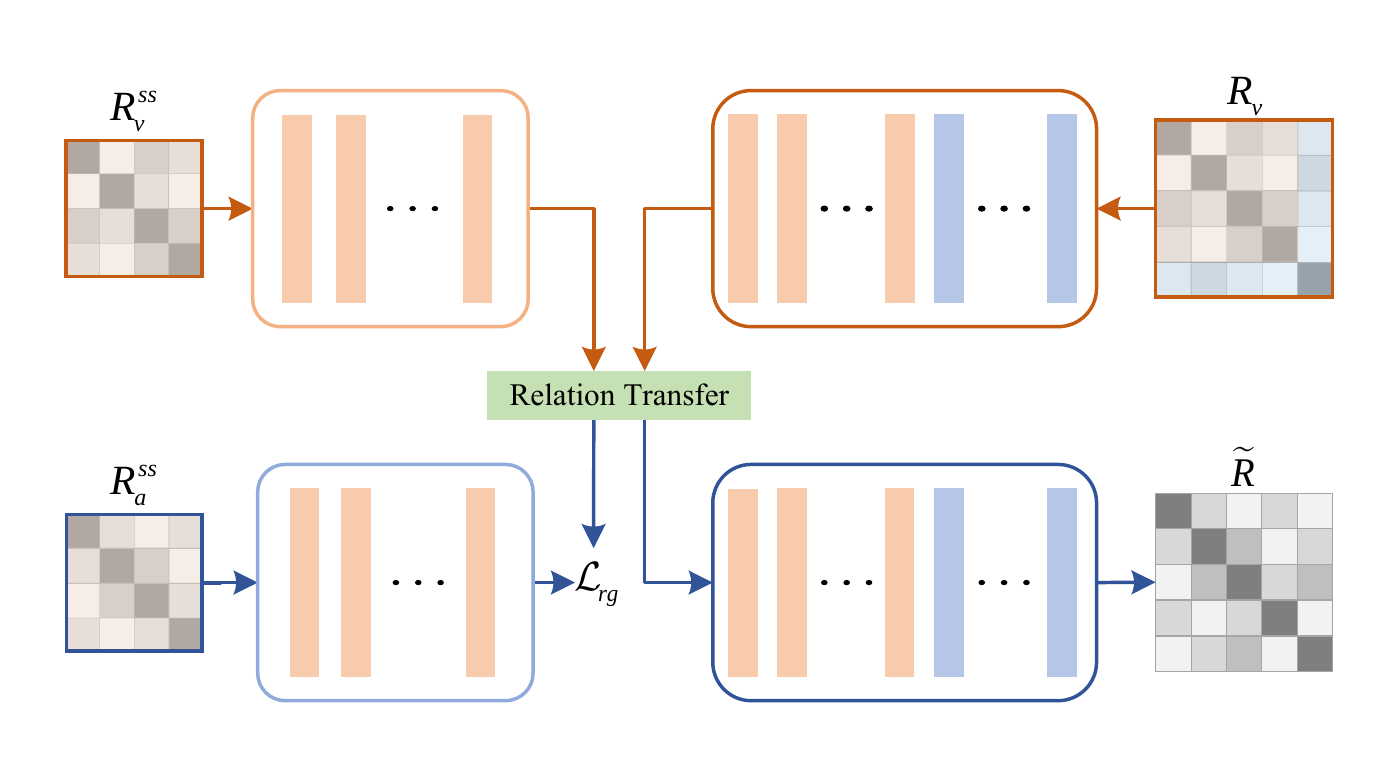}
	\end{center}
	\label{fig3}
	\caption{ Illustration of Relation Guidance (RG) strategy. There are two stages in RG strategy. Firstly, we train the Relation Transfer module with support-support pairs in visual and semantic modalities. Then, the entire visual relation map is rectified with the trained Relation Transfer module.
	}
\end{figure}

\textbf{Relation Guidance}. A regular way to guide the relation information across modal is regarding $A^v$ as the adjacency matrix $A^a$ in semantic graph. However, the visual feature embeddings are difficult to represent the corresponding samples correctly due to the unrelated information in images (e.g., background), while the semantic vectors are easier to discriminate. Therefore, $A^v$ is inappropriate for the relationships among samples, especially when the samples in each episode are scarce so that the distribution is incorrect. To obtain an appropriate adjacency matrix, we propose a Relation Guidance (RG) strategy to modify the relationships among visual samples with the guidance of support semantic vectors.

Specifically, we first obtain the relation maps $R_v$ and $R_a$ of two modalities. Each position of relation map is a relation vector representing the difference between two samples, which is calculated as follows:

\begin{equation}
r_{ij} = ( z_{i} - z_{j} )^2.
\end{equation}   
The relation map can be denoted by four parts as follows:

\begin{equation}
	R_v={
		\left( \begin{array}{cc}
		R_{v}^{ss} & R_{v}^{sq}\\
		R_{v}^{qs} & R_{v}^{qq}
		\end{array} 
		\right )},
\end{equation}

\begin{equation}
	R_a={
		\left( \begin{array}{cc}
		R_{a}^{ss} & 0 \\
		0 & 0
		\end{array} 
		\right )},
\end{equation}
where the relation vectors of the query-support and query-query pairs in semantic embeddings are 0. Thus, we design RG to employ the corresponding known parts to inference the unknown parts. Its core idea is to utilize the guidance of relations of support-support pairs in semantic graph to transfer $R_v$ to $\widetilde{R}$, which is more accurate to represent the relationships among different samples. Concretely, we set $R_v^{ss}$ as training samples and $R_a^{ss}$ as corresponding labels to train the relation transfer module $h(\cdot)$ with the mean square error (MSE) as loss function:

\begin{equation}
	\mathcal{L}_{rg}=\frac{1}{|R^{ss}|} \sum_{r_i \in R^{ss}} (\widetilde{r_i^v} - r_i^a)^2 ,
\end{equation}
where $\widetilde{r_i^v}=h(r_i^v)$. Then all relation vectors in $R_v$ will be fed into RG-module to obtain the modified $\widetilde{R}$:

\begin{equation}
	\widetilde{R} = h(R_v) = {
		\left( \begin{array}{cc}
		h(R_{v}^{ss}) & h(R_{v}^{sq})\\
		h(R_{v}^{qs}) & h(R_{v}^{qq})
		\end{array} 
		\right )}.
\end{equation}

Then the rectified adjacency matrix $\widetilde{A}$ can be acquired as follows:

\begin{equation}
	\widetilde{A}_{ij} = exp (-\frac{|\widetilde{R}_{ij}|}{\sigma^2}),
\end{equation}
where $|\widetilde{R}_{ij}|$ represents the distance between $z_i$ and $z_j$, which is the $l_1$-norm of the rectified relation vector.

\textbf{Graph Propagation}. With rectified adjacency matrix, both the visual graph and the semantic graph can be updated with graph propagation. Concretely, the matrix $\widetilde{A}$ is first symmetrically normalized as

\begin{equation}
	\widetilde{S} = \widetilde{D}^{-1/2} \widetilde{A} \widetilde{D}^{-1/2}  ,
\end{equation}
where $\widetilde{D}$ is the degree matrix of graph. Then, we follow the label propagation operation as in \cite{liu2019learning} and get the propagation matrix as

\begin{equation}
	\widetilde{P} = (1 - \alpha) (I - \alpha \widetilde{S})^{-1} ,
\end{equation}
where $\alpha \in (0,1) $ is a smoothing factor and $I$ is the identity matrix. Finally, the visual embeddings can be rectified as

\begin{equation}
	\widetilde{z^v} = \widetilde{P} \cdot z^v ,
\end{equation}
and the completed semantic embeddings can be obtained as

\begin{equation}
	\widetilde{z^a} = \widetilde{P} \cdot z^a .
\end{equation}

With guidance of the pseudo query semantic embeddings and the existing support semantic embeddings, the feature embeddings are modified to be more discriminative. Most importantly, the information asymmetries between support set and query set are reduced significantly.

\subsection{Feature Fusion and Classification}

After the modal-alternating propagation module, the information asymmetries are reduced significantly as the visual feature embeddings and semantic embeddings are both available for every sample. Since the information of the two modalities is complementary to enhance the features, we fuse embeddings by employing a convex combination to obtain more discriminative embeddings, which is denoted as:

\begin{equation}
	\widetilde{z} = \lambda \cdot \widetilde{z^v} + (1 - \lambda) \cdot \widetilde{z^a}.
\end{equation}
where $\lambda$ is a coefficient learned with a weight learner:

\begin{equation}
	\lambda = w( \widetilde{z^v} || \widetilde{z^a} ) ,
\end{equation}
where $w$ is an MLP and $||$ represents the concatenating operation. Then following the operation of Prototypical Networks \cite{Snell2017Prototypical}, we calculate the classification loss as follows:

\begin{equation}
	\mathcal{L}_{cls} = \frac{1}{|Q|} \sum_{i} \log p(y_i=c | x_i \in Q) ,
\end{equation}

\begin{equation}
	p(y_i=c | x_i \in Q) = \frac{ exp (-d(\widetilde{z}_i, \widetilde{p}_c)) }{ \sum_{j} exp(-d(\widetilde{z}_j, \widetilde{p}_c)) } ,
\end{equation}
where $d$ is the Euclidean distance, $p(y_i=c | x_i \in Q)$ is the probability that query sample $x_i$ belongs to class $c$ and $\widetilde{p}_c$ is the prototype of augmented features in class $c$.

\begin{equation}
	\widetilde{p}_c = \frac{1}{|S_c|} \sum_{ \widetilde{z}_i \in S_c } \widetilde{z}_i .
\end{equation}

Therefore, the total loss is

\begin{equation}
	\mathcal{L} = \mathcal{L}_{cls} + \mu \mathcal{L}_{rg},
\end{equation}
where $\mu$ is the weight coefficient of relation transfer loss.

\setcounter{table}{1}
\begin{table*}[t]
	\label{Table.2}
	\caption{\upshape  Few-shot classification accuracy on CUB with $\pm$ 95\% confidence intervals. $\dag$ denotes that the accuracies are reported in \cite{huang_attributes-guided_2021}.}
	\centering
	\setlength{\tabcolsep}{5mm}
	\renewcommand\arraystretch{1.2}
	\begin{tabular}{lcccc}
		\hline 
		\multirow{2}*{Methods}        & \multirow{2}*{Semantic}    & \multirow{2}*{Backbone}      & \multicolumn{2}{c}{Accuracy}         \\
		~                             & ~                          & ~	                          & 5-way 1-shot          & 5-way 5-shot      \\
		\hline
		MatchingNet \cite{vinyals2016matching}            & N      &ConvNet4     & 60.52 $\pm$ 0.88\%    & 75.29 $\pm$ 0.75\% \\
		ProtoNet \cite{Snell2017Prototypical}             & N      &ConvNet4     & 50.46 $\pm$ 0.88\%    & 76.39 $\pm$ 0.64\% \\
		RelationNet \cite{sung2018learning}               & N      &ConvNet4     & 62.34 $\pm$ 0.94\%    & 77.84 $\pm$ 0.68\% \\
		MAML \cite{finn2017model-agnostic}                & N      &ConvNet4     & 54.73 $\pm$ 0.97\%    & 75.75 $\pm$ 0.75\% \\
		ARML \cite{Yao2020Automated}                      & N      &ConvNet4     & 62.33 $\pm$ 1.47\%    & 73.34 $\pm$ 0.70\% \\
		\hline
		SoSN-ArL \cite{zhang_rethinking_2021}             & Y      &ConvNet4     & 50.62\%               & 65.87\% \\
		AM3$^\dag$ \cite{xing_adaptive_2019}              & Y      &ConvNet4     & 73.78 $\pm$ 0.28\%    & 81.39 $\pm$ 0.26\% \\
		AGAM \cite{huang_attributes-guided_2021}          & Y      &ConvNet4     & 75.87 $\pm$ 0.29\%    & 81.66 $\pm$ 0.25\% \\
		\hline
		\textbf{MAP-Net (Ours)}                           & Y      &ConvNet4     & \textbf{80.92 $\pm$ 0.21\%}    & \textbf{85.88 $\pm$ 0.17\%} \\
		\hline
		MatchingNet$^\dag$ \cite{vinyals2016matching}     & N      &ResNet12     & 60.96 $\pm$ 0.35\%    & 77.31 $\pm$ 0.25\% \\
		ProtoNet \cite{Snell2017Prototypical}             & N      &ResNet12     & 68.8\%                & 76.4\% \\
		RelationNet$^\dag$ \cite{sung2018learning}        & N      &ResNet12     & 60.21 $\pm$ 0.35\%    & 80.18 $\pm$ 0.25\% \\
		MAML \cite{finn2017model-agnostic}                & N      &ResNet18     & 69.96 $\pm$ 1.01\%    & 82.70 $\pm$ 0.65\% \\
		TADAM \cite{oreshkin2018tadam}                    & N      &ResNet12     & 69.2\%                & 78.6\% \\
		FEAT \cite{ye2020few}                             & N      &ResNet12     & 68.87 $\pm$ 0.22\%    & 82.90 $\pm$ 0.15\% \\
		AFHN \cite{li2020adversarial}                     & N      &ResNet18     & 70.53 $\pm$ 1.01\%    & 83.95 $\pm$ 0.63\% \\
		\hline
		Comp. \cite{tokmakov_learning_2019}               & Y      &ResNet10     & 53.6\%                & 74.6\% \\
		AM3 \cite{xing_adaptive_2019}                     & Y      &ResNet12     & 73.6\%                & 79.9\% \\
		Dual TriNet \cite{chen2019multi}                  & Y      &ResNet12     & 69.61 $\pm$ 0.46\%    & 84.10 $\pm$ 0.35\% \\
		Multi-Sem. \cite{schwartz_baby_2019}              & Y      &DenseNet121  & 76.1\%                & 82.9\% \\
		AGAM \cite{huang_attributes-guided_2021}          & Y      &ResNet12     & 79.58 $\pm$ 0.23\%    & 87.17 $\pm$ 0.23\% \\
		\hline
		\textbf{MAP-Net (Ours)}                           & Y      &ResNet12     & \textbf{82.45 $\pm$ 0.23\%}    & \textbf{88.30 $\pm$ 0.17\%} \\
		\hline

	\end{tabular}
\end{table*}

\section{Experiments}
\subsection{Experiment Setup}

\setcounter{table}{0}
\begin{table}[htbp]
	\label{Table.1}
	\caption{\upshape   Information of Datasets.} 
	\centering
	\setlength{\tabcolsep}{0.5mm}
	\renewcommand\arraystretch{1.2}
	\begin{tabular}{ccccc}
		\hline 
		Datasets   &  Training set   &   Validation set   & Testing set    &  Semantics   \\
		\hline
		CUB-200-2011 \cite{wah2011caltech}     & 100   & 50    & 50    &  Attributes           \\
		SUN Attribute \cite{patterson2014sun}    & 580   & 65    & 72    &  Attributes           \\
		Flowers 102 \cite{nilsback2008automated}     &60    & 20    & 22    &  Text-descriptions    \\
		
		\hline
	\end{tabular}
\end{table}

\textbf{Datasets}. 
We conduct the experiments on three benchmark datasets with semantic information, i.e., attributes or text descriptions: Caltech-UCSD-Birds 200-2011 (CUB) \cite{wah2011caltech}, SUN Attribute Database (SUN) \cite{patterson2014sun} and Oxford 102 Flower (Flower) \cite{nilsback2008automated}. The details about datasets are shown in Table \uppercase\expandafter{\romannumeral1}. CUB is a fine-grained dataset of bird species that consists of 11788 images of 200 categories and 312 attributes for each class. We follow the split in \cite{chen2019a}, which selects 100, 50, 50 classes for training, validation and testing respectively. SUN contains 14340 scene images of 717 classes with 102 attributes. Following \cite{huang_attributes-guided_2021}, 580, 65, 72 classes will be employed for training, validation and testing. Flower is also a fine-grained dataset with 102 classes of flower species, where the number of images is varied from 40$ \sim $258 in each class. Each image has a text description in 1024 dimensions. 60, 20, 22 classes are used for training, validation and testing. All images in these datasets are resized to 84$\times$84 for fair comparisons.

\setcounter{table}{2}
\begin{table}[htbp]
	\label{Table.3}
	\caption{\upshape  Few-shot classification accuracy of SUN with $\pm$  95\% confidence intervals. $\dag$ denotes that the accuracies are reported in \cite{huang_attributes-guided_2021}.}
	\centering
	\setlength{\tabcolsep}{2mm}
	\renewcommand\arraystretch{1.2}
	\begin{tabular}{lccc}
		\hline
		\multirow{2}*{Methods}             & \multirow{2}*{Backbone}      & \multicolumn{2}{c}{Accuracy}         \\
		~                                  & ~	                          & 5-way 1-shot          & 5-way 5-shot      \\
		\hline
		MatchingNet$^\dag$ \cite{vinyals2016matching}             &ConvNet4      & 55.72 $\pm$ 0.40\%  & 76.59 $\pm$ 0.21\% \\
		ProtoNet$^\dag$ \cite{Snell2017Prototypical}              &ConvNet4      & 57.76 $\pm$ 0.29\%  & 79.27 $\pm$ 0.19\% \\
		RelationNet$^\dag$ \cite{sung2018learning}                &ConvNet4      & 49.58 $\pm$ 0.35\%  & 76.21 $\pm$ 0.19\% \\
		\hline
		Comp. \cite{tokmakov_learning_2019}                       &ResNet10      & 45.9\%              & 67.1\% \\
		AM3$^\dag$ \cite{xing_adaptive_2019}                      &ConvNet4      & 62.79 $\pm$ 0.32\%  & 79.69 $\pm$ 0.23\% \\
		AGAM \cite{huang_attributes-guided_2021}                  &ConvNet4      & 65.15 $\pm$ 0.31\%  & 80.08 $\pm$ 0.21\% \\
		\hline
		\textbf{MAP-Net (Ours)}                                   &ConvNet4      & \textbf{67.73 $\pm$ 0.30\%} & 	\textbf{80.30  $\pm$ 0.21\%} \\
		\hline
	\end{tabular}
\end{table}

\begin{table}[htbp]
	\label{Table.4}
	\caption{\upshape  Few-shot classification accuracy of Flowers with $\pm$  95\% confidence intervals. $\ast$ denotes that it is our implementation. $\dag$ denotes that the accuracies are reported in \cite{zhang_rethinking_2021}.}
	\centering
	\setlength{\tabcolsep}{2mm}
	\renewcommand\arraystretch{1.2}
	\begin{tabular}{lccc}
		\hline
		\multirow{2}*{Methods}             & \multirow{2}*{Backbone}      & \multicolumn{2}{c}{Accuracy}         \\
		~                                  & ~	                          & 5-way 1-shot          & 5-way 5-shot      \\
		\hline
		ProtoNet$^\dag$ \cite{Snell2017Prototypical}              &ConvNet4      & 62.81\%            & 82.11\% \\
		RelationNet$^\dag$ \cite{sung2018learning}                &ConvNet4      & 68.26\%            & 80.94\% \\
		\hline
		AM3$^\ast$ \cite{xing_adaptive_2019}                      &ConvNet4      & 74.00 $\pm$ 0.29\%  & 88.67 $\pm$ 0.18\% \\
		AGAM$^\ast$ \cite{huang_attributes-guided_2021}           &ConvNet4      & 73.27 $\pm$ 0.29\%  & 89.49 $\pm$ 0.17\% \\
		SoSN-ArL \cite{zhang_rethinking_2021}              &ConvNet4      & 76.21\%              & 88.39\% \\
		\hline
		\textbf{MAP-Net (Ours)}                            &ConvNet4      & \textbf{77.41 $\pm$ 0.28\%} & 	\textbf{90.63 $\pm$ 0.16\%} \\
		\hline
	\end{tabular}
\end{table}

\begin{table*}[t]
	\label{Table.5}
	\caption{\upshape  Ablation results of main components. Notations: ‘VP’ - Visual Propagation, ‘SP’ - Semantic Propagation, ‘RG’ - Relation Guidance.}
	\centering
	\setlength{\tabcolsep}{3mm}
	\renewcommand\arraystretch{1.2}
	\begin{tabular}{ccccccccc}
		\hline
		\multirow{2}*{VP} & \multirow{2}*{SP}  & \multirow{2}*{RG} & \multicolumn{2}{c}{CUB}       & \multicolumn{2}{c}{SUN}   & \multicolumn{2}{c}{Flowers}      \\
		~ & ~  & ~ & 5-way 1-shot  & 5-way 5-shot  & 5-way 1-shot  & 5-way 5-shot  & 5-way 1-shot  & 5-way 5-shot      \\
		\hline
		  &  &                                 &  75.30 $\pm$ 0.25\% & 80.28 $\pm$ 0.22\%   & 63.11 $\pm$ 0.30\%  & 78.91 $\pm$ 0.21\%  & 74.78 $\pm$ 0.28\%  & 88.81 $\pm$ 0.17\% \\
		\checkmark &  &                        &  78.14 $\pm$ 0.24\% & 81.42 $\pm$ 0.22\%   & 59.12 $\pm$ 0.31\%  & 76.51 $\pm$ 0.25\%  & 74.97 $\pm$ 0.28\%  & 89.76 $\pm$ 0.17\% \\
		  & \checkmark  &                      &  79.53 $\pm$ 0.23\% & 84.08 $\pm$ 0.19\%   & 66.90 $\pm$ 0.29\%  & 79.82 $\pm$ 0.23\%  & 76.14 $\pm$ 0.28\%  & 90.19 $\pm$ 0.17\% \\
		\checkmark & \checkmark  &             &  80.03 $\pm$ 0.23\% & 84.52 $\pm$ 0.19\%   & 66.83 $\pm$ 0.29\%  & 79.70 $\pm$ 0.23\%  & 76.28 $\pm$ 0.28\%  & 89.78 $\pm$ 0.17\% \\
		  & \checkmark & \checkmark            &  79.98 $\pm$ 0.22\% & 85.15 $\pm$ 0.19\%   & 67.35 $\pm$ 0.30\%  & 79.88 $\pm$ 0.24\%  & 76.68 $\pm$ 0.29\%  & 90.38 $\pm$ 0.16\% \\
		\checkmark & \checkmark &  \checkmark  &  \textbf{80.92 $\pm$ 0.21\%} & \textbf{85.88 $\pm$ 0.18\%}   & \textbf{67.73 $\pm$ 0.30\%}  & \textbf{80.30 $\pm$ 0.21\%}  & \textbf{77.41 $\pm$ 0.28\%}  & \textbf{90.63 $\pm$ 0.16\%} \\

		\hline
	\end{tabular}
\end{table*}

\textbf{Experimental Settings}. We conduct experiments on 5-way 1-shot and 5-way 5-shot settings, and 15 query samples are employed for both meta-training and meta-testing in each episode. The average accuracy (\%) and the corresponding 95\% confidence interval over the 5000 episodes are reported to express the performance. To make a fair comparison, our method is trained in the inductive setting. This is to say, we apply only one query sample to build corresponding graph at a time.

\textbf{Implementation Details}. We utilize two popular convolution networks ConvNet-4 \cite{vinyals2016matching} and ResNet-12 \cite{chen2020new} as backbones to extract the visual features of images, while the semantic embeddings are encoded with a Multi-Layer Perceptron (MLP). In the meta-training stage, we train the model with 60 epochs with 1000 episodes and 600 episodes for validation per epoch both in 1-shot and 5-shot settings. We randomly select K support samples and 15 query samples in each episode. The Adam \cite{kingma2014adam} optimizer is utilized with the initial learning rate 0.001 which will be reduced by 0.1 in every 15 epochs. The batch size is 5 for ConvNet-4 and 1 for ResNet-12. For MAP-Net, the smoothing factor $\alpha$ is set to 0.2 and weight coefficient $\mu$ is 1 for CUB and 0.1 for SUN and Flower.

\subsection{Comparison with State-of-the-art Methods}

We choose fourteen popular meta-learning based methods as competitors, including both nonsemantic-based methods and semantic-based methods:
	
	
	

	
	
	

(1) Nonsemantic-based methods.
\begin{itemize}
	\item Metric-based methods: Matching Networks \cite{vinyals2016matching}, Prototypical Networks \cite{Snell2017Prototypical}, Relation Networks \cite{sung2018learning}, TADAM \cite{oreshkin2018tadam} and FEAT \cite{ye2020few}.
	\item Optimization-based methods: MAML \cite{finn2017model-agnostic}, ARML \cite{Yao2020Automated}.
	\item Data-augmentation-based methods: AFHN \cite{li2020adversarial}.
\end{itemize}

(2) Semantic-based methods.
\begin{itemize}
	\item Guiding embeddings with semantics: AM3 \cite{xing_adaptive_2019}, Multi-Semantics \cite{schwartz_baby_2019}, Comp. \cite{tokmakov_learning_2019}, SoSN-ArL \cite{zhang_rethinking_2021}, AGAM \cite{huang_attributes-guided_2021}.
	\item Generating embeddings with semantics: Dual TriNet \cite{chen2019multi}.
\end{itemize}
	
We compare our MAP-Net with these methods with both ConvNet-4 and ResNet-12 on CUB. Since there is no method in previous conduct the experiments with ResNet-12 on SUN and Flower, we compare our method with these methods on them only with the ConvNet-4. Table \uppercase\expandafter{\romannumeral2}, \uppercase\expandafter{\romannumeral3}, \uppercase\expandafter{\romannumeral4} show the results on three benchmark datasets respectively.

It can be observed that our method outperforms all methods on three datasets in cases of ConvNet-4 and ResNet-12 backbones. Particularly, our MAP-Net achieves 5.05\% and 4.22\% performance gains on CUB, 2.58\% and 0.22\% on SUN, and 1.2\% and 1.14\% on Flower with the backbone ConvNet-4 against the second-best method on both 1-shot and 5-shot settings. For ResNet-12 on CUB, MAP-Net also obtains the best performance on both 1-shot and 5-shot settings, which outperforms the second-best approach AGAM \cite{huang_attributes-guided_2021} in 2.87\% and 1.13\% respectively. 

Besides, we have the following three observations. 
(1) The improvements are more significant on 1-shot setting than those on 5-shot setting. It proves that the auxiliary semantic information is more beneficial in the occasion with extremely few visual samples. 
(2) The improvements with ResNet-12 are inferior to those with ConvNet-4. It demonstrates that the complex feature encoder suppresses the effectiveness of semantic information. When the feature encoder extracts more discriminative embeddings, the balance between visual and semantic modalities is broken due to that the information carried from semantic vectors is finite. 
(3) It is observed that the performance is improved slightly on 5-shot setting of SUN. We suppose the main reason is that the attribute vectors of SUN are too sparse that the effectiveness is suppressed when the visual information increasing. 

\begin{figure*}[t]
	\tiny
	\begin{center}
		\includegraphics[width=0.9\textwidth]{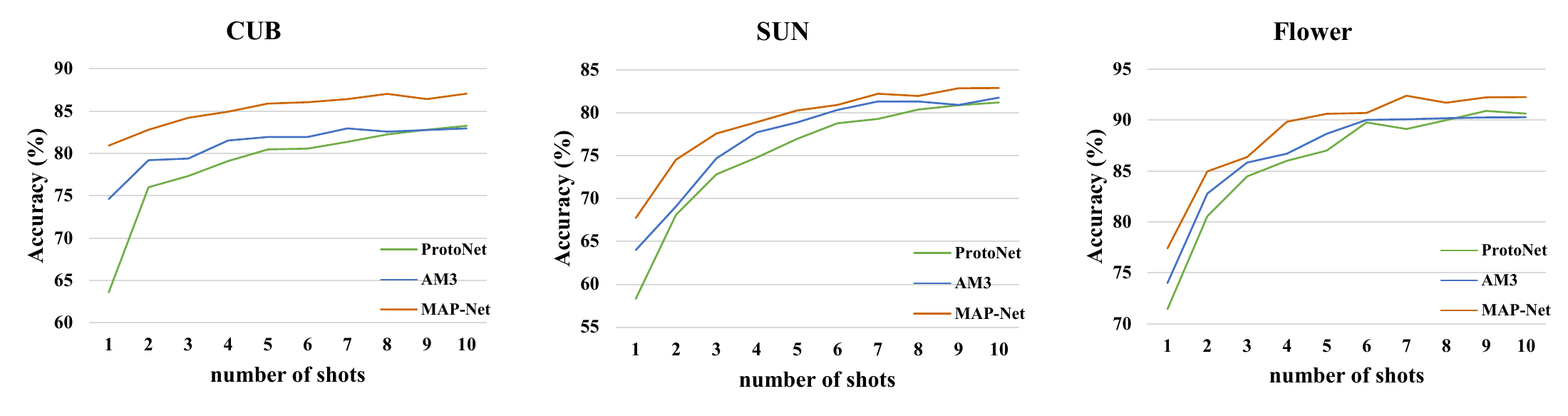}
	\end{center}
	\label{fig4}
	\caption{Comparison of our method and two baseline methods with different number of shots on CUB, SUN, Flower.
	}
\end{figure*}

\begin{figure*}[t]
	\tiny
	\begin{center}
		\includegraphics[width=0.9\textwidth]{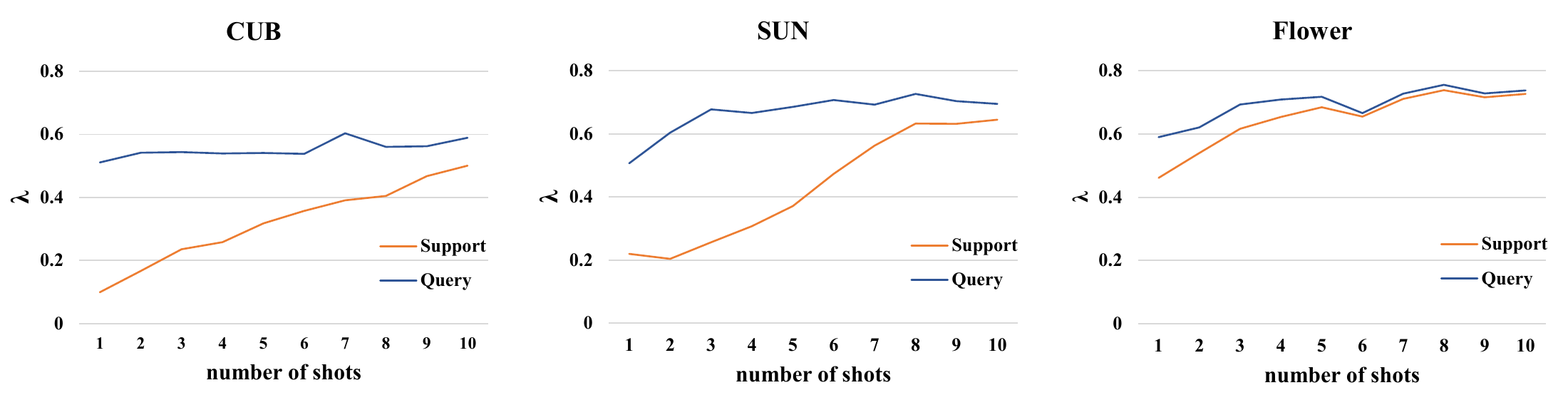}
	\end{center}
	\label{fig5}
	\caption{Average value of $\lambda$ for support samples and query samples respectively with different number of shots learned on CUB, SUN, Flower.
	}
\end{figure*}

\subsection{Ablation Studies} 
We conduct ablation studies to prove the effectiveness of the main components in MAP-Net on three datasets, which are shown in Table \uppercase\expandafter{\romannumeral5}. We set the model without these components as the baseline, which is similar to AM3 \cite{xing_adaptive_2019}. The only difference is that AM3 combines the visual prototypes with the class semantic embeddings, while our baseline combines support visual features with their semantic embeddings.

It is observed that our method with all components obtains the best performance on three datasets. Both visual graph propagation and semantic graph propagation are beneficial in MAP-Net, especially the propagation in semantic modality. For example, compared with the baseline, the visual propagation brings 2.84\% and 1.14\% improvements on 1-shot and 5-shot on CUB, while the semantic propagation brings 4.23\% and 3.8\% improvements. For SUN and Flower, the semantic propagation also improves the accuracy by a large margin. It demonstrates that the information asymmetries between support and query samples are reduced significantly with semantic propagation. However, the visual propagation does not bring significant gains, and the performance is even damaged on SUN. We suppose the reason is that the information propagated from neighbor nodes in the visual graph might be useless or even harmful since the relationships among samples are not quite accurate in low-data scenarios. 

\begin{table}[htbp]
	\label{Table.6}
	\caption{\upshape  Results of different guidance methods on CUB. Notations: ‘IC’ - Instance Constraint, ‘RC’ - Relation Constraint, ‘RG’ - Relation Guidance.}
	\centering
	\setlength{\tabcolsep}{1.3mm}
	\renewcommand\arraystretch{1.2}
	\begin{tabular}{lccc}
		\hline
		\multirow{2}*{Methods}             & \multirow{2}*{Backbone}      & \multicolumn{2}{c}{Accuracy}         \\
		~                                  & ~	                          & 5-way 1-shot          & 5-way 5-shot      \\
		\hline
		MAP-Net (w/o guidance)             &ConvNet4      & 80.03 $\pm$ 0.23\%            & 84.52 $\pm$ 0.19\% \\
		MAP-Net (w IC)                     &ConvNet4      & 78.62 $\pm$ 0.24\%            & 82.50 $\pm$ 0.21\% \\
		MAP-Net (w RC)                     &ConvNet4      & 79.13 $\pm$ 0.24\%            & 82.65 $\pm$ 0.21\% \\
		MAP-Net (w RG)                     &ConvNet4      & \textbf{80.92 $\pm$ 0.21\%}   & \textbf{85.88 $\pm$ 0.17\%} \\
		
		\hline
	\end{tabular}
\end{table}

For the same reason, the performance is also limited when utilizing the visual propagation and the semantic propagation simultaneously. By contrast, this issue is alleviated with Relation Guidance strategy, which brings about 1\% on all settings. It indicates that with the guidance of semantic embeddings with accurate distribution, the rectified relation map represents the relationships among samples more correctly to propagate discriminative pseudo-semantic embeddings. Moreover, the adverse impact of visual propagation is also reduced.

\subsection{Further Analysis}

\textbf{Impact of Different Guidance Methods}. 
In order to validate the efficiency of the Relation Guidance strategy, we design some other methods to guide the visual embeddings with semantic information. In this section, we compare the results on CUB with different guidance methods: Instance Constraint (IC) – constraining every cross-modal representation of support samples directly; Relation Constraint (RC) – constraining the corresponding relation vectors of support-support pairs in two modalities; and our proposed Relation Guidance (RG) method. The results are reported in Table \uppercase\expandafter{\romannumeral6}.

We could observe that both IC and RC hurt the performance, which indicates that directly constraining cross-modal embeddings is inappropriate in few-shot learning. By contrast, MAP-Net with RG achieves promising performance. It demonstrated that the information propagated among nodes is more accurate with the Relation Guidance strategy. The relative information is more effective to be utilized for alleviating the cross-modal discrepancy. 

\begin{figure*}[t]
	\begin{center}
		\includegraphics[width=0.9\textwidth]{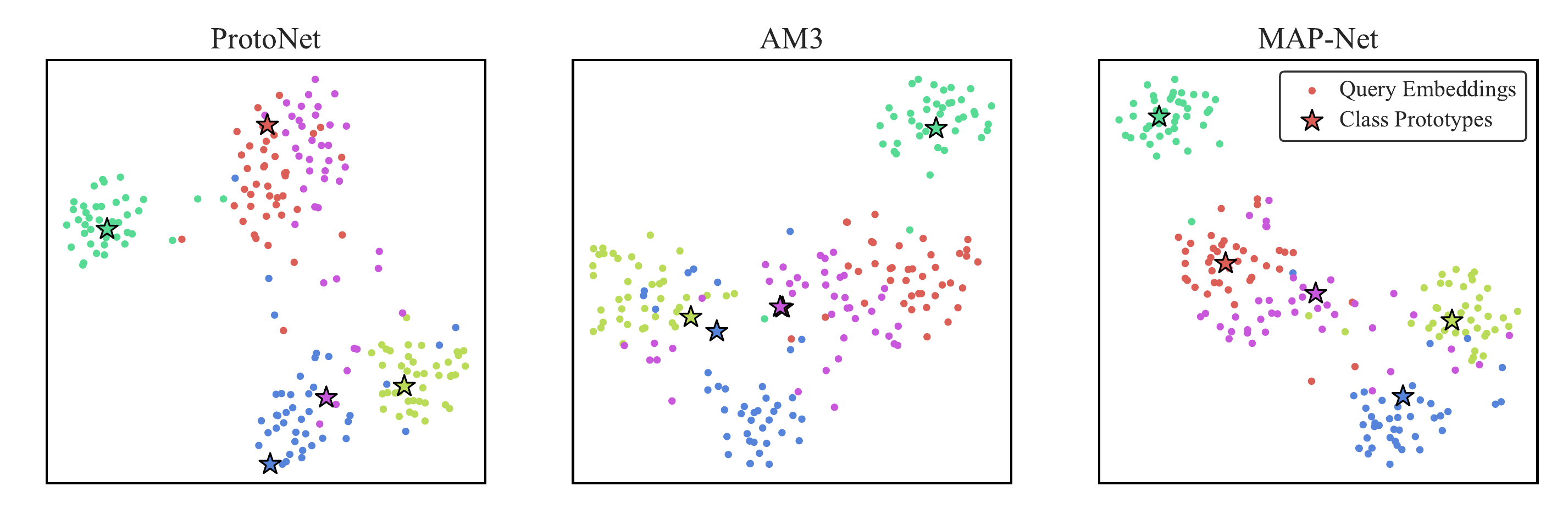}
	\end{center}
	\label{fig6}
	\caption{ The t-SNE visualization of the embeddings learned by ProtoNet, AM3, MAP-Net on CUB respectively.
	}
\end{figure*}

\textbf{Analysis of Information Asymmetries}.
Figure 4 shows the accuracies of our method and two baseline methods (AM3 and ProtoNet) on three datasets on 1-10 shot settings. It can be observed that the improvement of AM3 compared with ProtoNet is reduced when the number of shots increased. The ProtoNet even surpasses AM3 with 10-shot setting on CUB and Flower datasets. By contrast, our method outperforms these two baselines whatever the number of shots is. We interpret the result in two aspects. Firstly, the semantic information is more essential in the scenario with extremely few visual samples (e.g., 1-shot), while the abundant visual information (e.g., 10-shot) can suppress the effectiveness of semantic. Secondly, more supervision information reflects more accurate relationships among different samples which are beneficial to generate the pseudo-semantic embeddings of query samples to reduce the information asymmetries. Therefore, our method also improves the performance as the number of shots increases.

Figure 5 shows the relationship between the weight coefficient $\lambda$ and the number of shots. It is observed that the coefficient $\lambda$ is larger as the number of shots increases as analyzed in \cite{xing_adaptive_2019}. Moreover, $\lambda$ for query samples is larger than that for support samples. We consider the reason is that the pseudo-semantic embeddings of query samples propagated from adjacent support samples cannot represent the real semantic information thoroughly, though they supplement the missing information to some extent, which is worthy of further study in the future. We can also observe that with the increase of shots, the gap between the values of $\lambda$ for support samples and query samples is decreasing. It means that the pseudo-semantic of query samples and real semantic of support samples are more symmetric with abundant relationship information constructed with more support samples, which also proves the effectiveness of modal-alternating propagation.

The difference among three datasets expresses the different representativeness of semantic information to corresponding samples in different datasets. The semantic information used in CUB is more representative since its weight is large, almost 0.9, on semantic embeddings, while those on SUN and Flower are relatively small.

\textbf{Visualization Analysis}.
To prove the effectiveness of our method, we visualize the embedding space on CUB dataset by the t-SNE approach. It is observed in Fig. 6 that the embeddings obtained from AM3 and MAP-Net are more separable than Prototypical Networks with the auxiliary of semantic information. As for ProtoNet, one prototype acquired by averaging support samples is even far from the corresponding query samples, which extremely affects the classification results. Moreover, although the clusters of AM3 are relatively scattered, there is a large distance between the query embeddings and their prototypes, which is caused by the information asymmetries due to the lack of query semantic information. And two prototypes even overlap with each other. By contrast, the query embeddings and their prototypes of MAP-Net are more consistent with the generated pseudo-semantic embeddings.

\section{Conclusion}
To address the information asymmetry problem when utilizing the auxiliary semantic information in few-shot learning, we have proposed a Modal-Alternating Propagation Network (MAP-Net) to supplement the lack of query semantics. The MAP-Net generates the query semantics and modifies the feature embeddings by employing the relationships to alternatingly propagate graphs in two modalities. Furthermore, with the proposed flexible Relation Guidance (RG) strategy, the large discrepancy between different modalities is reduced significantly. The relation vectors are more accurate to represent the true relationships among samples with the guidance of semantics. Extensive experiments on three datasets with semantic information have demonstrated the effectiveness of our proposed MAP-Net.

\bibliographystyle{ieeetr}
\bibliography{ref}

\begin{IEEEbiography}[{\includegraphics[width=1in,height=1.25in,clip,keepaspectratio]{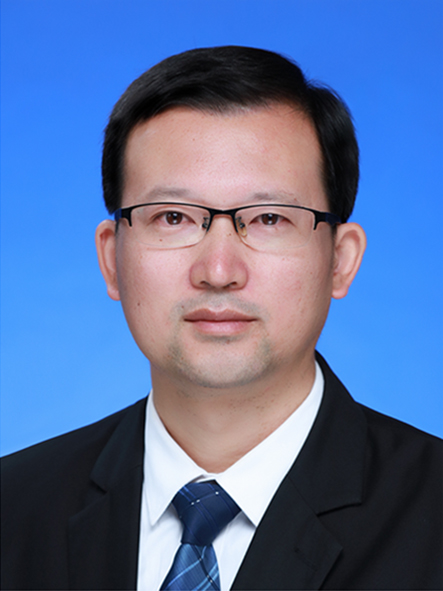}}]{Zhong Ji}
received the Ph.D. degree in signal and information processing from Tianjin University, Tianjin, China, in 2008. He is currently a Professor with the School of Electrical and Information Engineering, Tianjin University. He has authored over 100 scientific papers. His current research interests include multimedia understanding, zero/few-shot leanring, cross-modal analysis, and video summarization.
\end{IEEEbiography}

\begin{IEEEbiography}[{\includegraphics[width=1in,height=1.25in,clip,keepaspectratio]{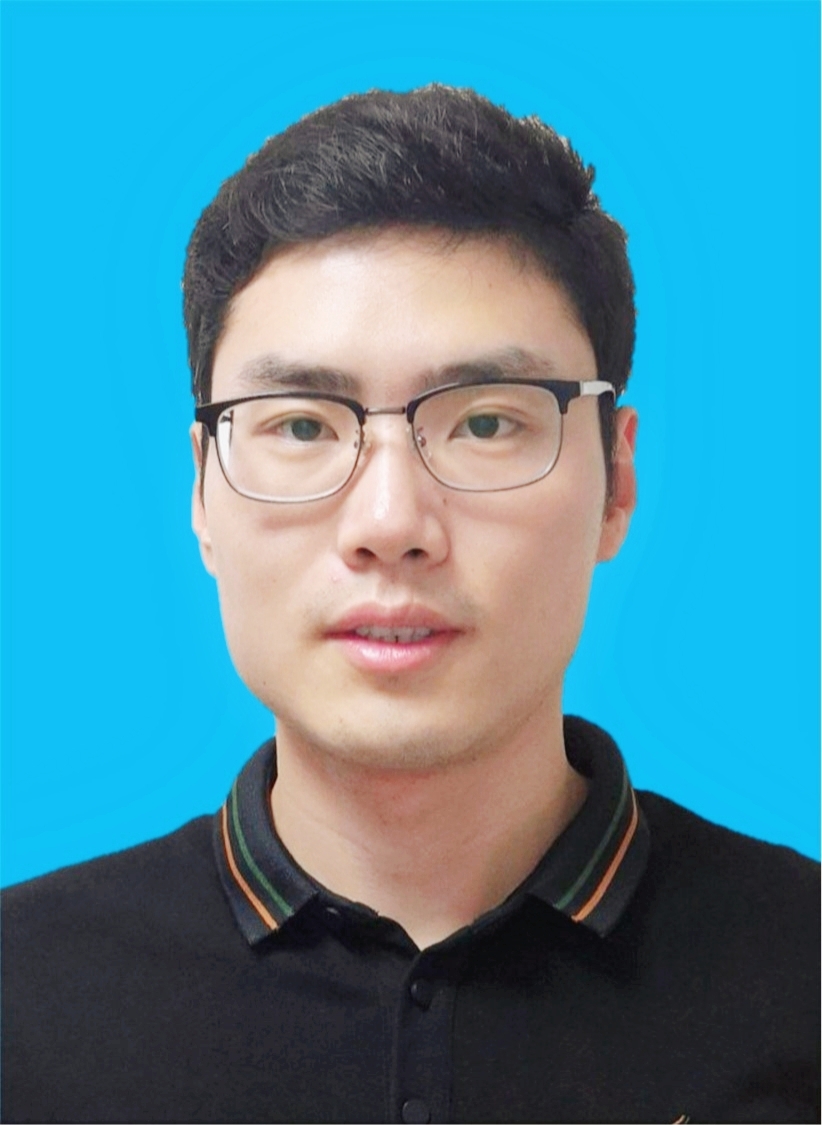}}]{Zhishen Hou}
received the B.S. degree in electronic and information engineering from Tianjin University, Tianjin, China, in 2020. He is currently pursuing a M.S. degree in the School of Electrical and Information Engineering, Tianjin University. His research interests include few-shot learning and computer vision.
\end{IEEEbiography}

\begin{IEEEbiography}[{\includegraphics[width=1in,height=1.25in,clip,keepaspectratio]{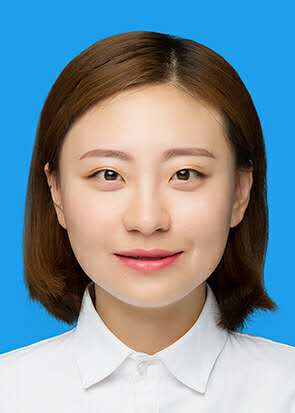}}]{Xiyao Liu}
received the B.S. degree in telecommunication engineering from Tianjin University, Tianjin, China, in 2015. She is currently pursuing a Ph.D. degree in the School of Electrical and Information Engineering, Tianjin University. Her research interests include few-shot learning, human-object interaction, and computer vision.
\end{IEEEbiography}

\begin{IEEEbiography}[{\includegraphics[width=1in,height=1.25in,clip,keepaspectratio]{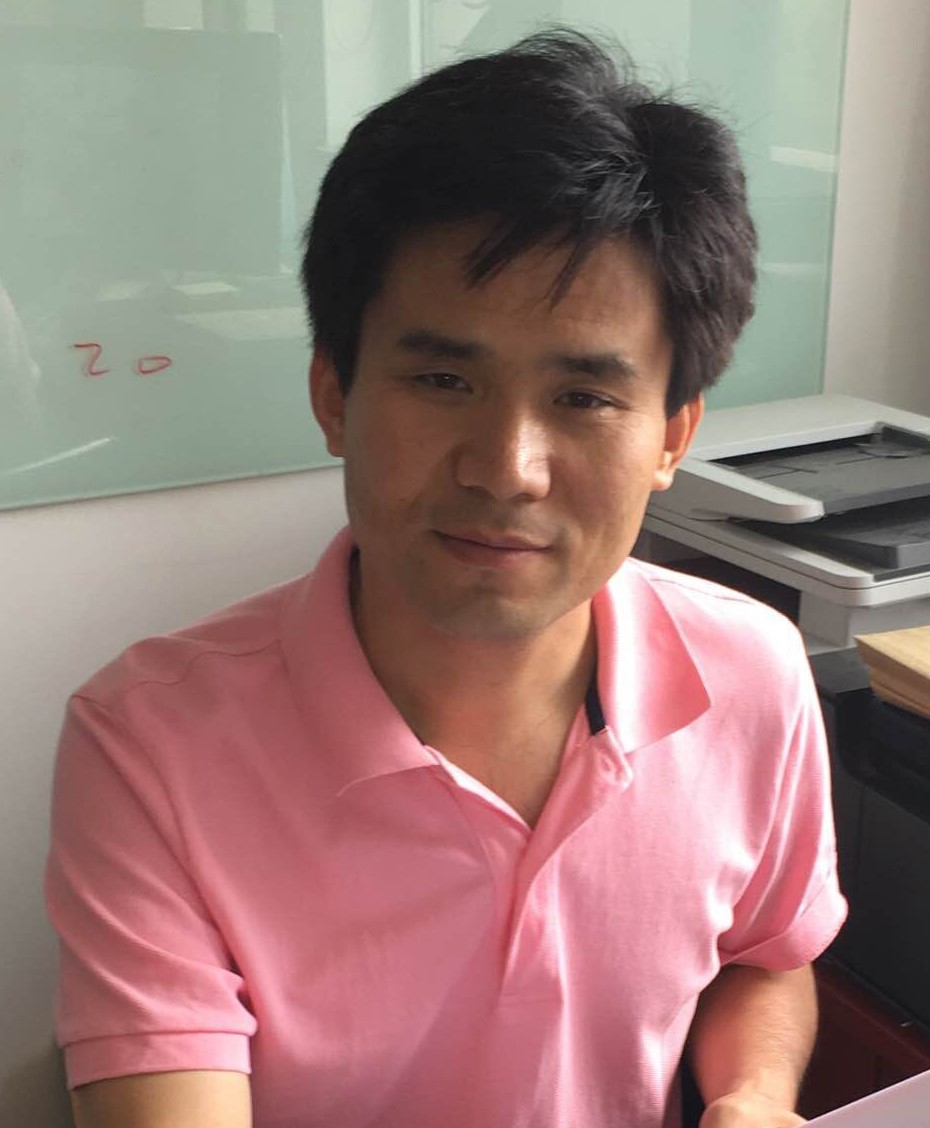}}]{Yanwei Pang}
received the Ph.D. degree in electronic engineering from the University of Science and Technology of China, Hefei, China, in 2004. He is currently a Professor with the School of Electrical and Information Engineering, Tianjin University, Tianjin, China. He has authored over 120 scientific papers. His current research interests include object detection and recognition, vision in bad weather, and computer vision.

\end{IEEEbiography}

\begin{IEEEbiography}[{\includegraphics[width=1in,height=1.25in,clip,keepaspectratio]{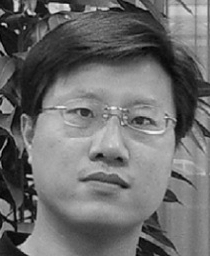}}]{Jungong Han}
 is currently a Full Professor and Chair in Computer Science at Aberystwyth University, UK. His research interests span the fields of video analysis, computer vision and applied machine learning. He has published over 180 papers, including 50+ IEEE Trans and 40+ A* conference papers.
\end{IEEEbiography}


\end{document}